\def\year{2021}\relax
\documentclass[letterpaper]{article} % DO NOT CHANGE THIS
\usepackage{aaai21}  % DO NOT CHANGE THIS
\usepackage{multirow}
\usepackage{array}
\usepackage{makecell}
\usepackage{enumitem}
\usepackage{bm}
\usepackage{times}  % DO NOT CHANGE THIS
\usepackage{helvet} % DO NOT CHANGE THIS
\usepackage{courier}  % DO NOT CHANGE THIS
\usepackage[hyphens]{url}  % DO NOT CHANGE THIS
\usepackage{graphicx} % DO NOT CHANGE THIS

\usepackage{natbib}  % DO NOT CHANGE THIS AND DO NOT ADD ANY OPTIONS TO IT
\usepackage{caption} % DO NOT CHANGE THIS AND DO NOT ADD ANY OPTIONS TO IT
\title{Natural Language for Human-Robot Collaboration: \\ Problems Beyond Language Grounding}
\author{
    Seth Pate, Wei Xu, Ziyi Yang,
    Maxwell Love, Siddarth Ganguri,
    Lawson L.S. Wong
    \\
}
\affiliations{
    Khoury College of Computer Sciences, Northeastern University\\
  
    \{ pate.s , xu.wei3 , yang.ziyi2 , love.m , ganguri.s \}@northeastern.edu , lsw@ccs.neu.edu

}

\begin{document}

\maketitle

\begin{abstract}

To enable robots to instruct humans in collaborations, we identify several aspects of language processing that are not commonly studied in this context.
These include location, planning, and generation.
We suggest evaluations for each task, offer baselines for simple methods,
and close by discussing challenges and opportunities in studying language for collaboration.

\end{abstract}

\section{Significance}

Humans generally want robots to do as they are told -- this is, indeed, the second of Asimov's laws.
But if a robot can solve a problem better than we can, perhaps it should be telling us what to do.

Most language research in robotics has been on language \textbf{grounding}, in which a robot translates human instructions into actions \citep{anderson2018vision,anderson2018evaluation,fried2018speaker,majumdar2020improving,li2021improving,wang2019reinforced}.
We take the reverse approach, identifying language skills with which robots may advise humans: \textbf{locating}, \textbf{planning}, and \textbf{speaking}.
Developing these understudied skills will improve robot grounding, but it will also make for better robot collaborators.

In collaboration, each partner brings its own abilities.
When robots are used for their strength, speed, and endurance, they need only understand language to follow instructions.
But as robots get better at problem solving, they can offer us more than obedience.

For example, in industrial contexts like manufacturing and shipping, robots often operate without much language ability.
But if an embodied agent could generate language, it could use its extensive memory, planning ability, and knowledge of its environment to provide its human teammate with reminders and instructions.
These instructions are often helpful even if imperfect.

In this paper, we define, describe, and summarize literature on the language skills robots need to advise us.
We set the collaboration problem in the context of navigation, suggest a common environment, and provide baselines with simple methods.
We close with a discussion of the challenges and opportunities in creating robots that can give instructions as well as take them.

\section{General Problem Definition and Related Work}

\begin{figure*}
\centering
\includegraphics[width=\linewidth]{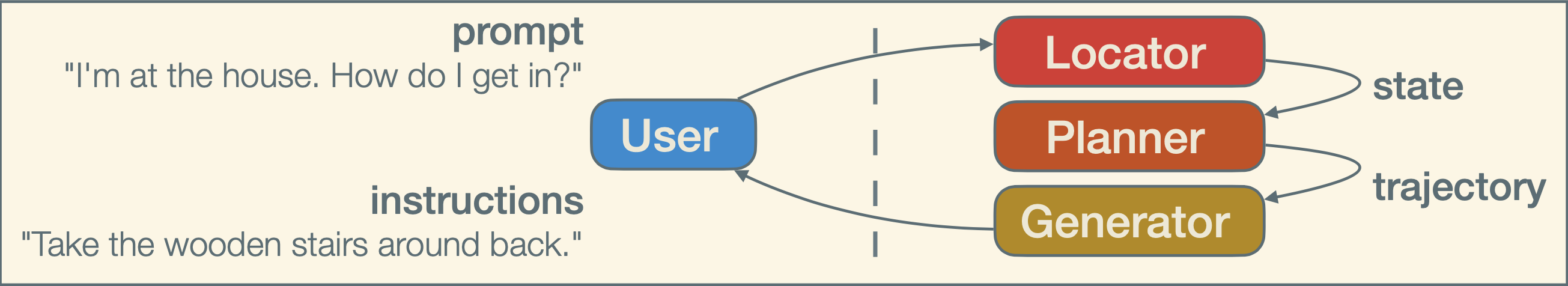}
\caption{The collaboration problem and its tasks as a cycle.}
\label{fig:rv}
\end{figure*}

For sake of generality, we refer to our robot as an \textbf{agent}, and its human partner as a \textbf{user}.
To give instructions, an agent needs several skills:

\begin{itemize}[leftmargin=*]
    \item \textbf{Location:} Interpret user's \textbf{language}, $\boldsymbol{l}$, to recognize the user's \textbf{current state} $s$ and \textbf{goal state} $g$.
    \item \textbf{Planning:} Find a \textbf{trajectory} $\boldsymbol{x}$ to bring the user from $s$ to $g$.
    \item \textbf{Generation:} Draft natural language \textbf{instructions} $\boldsymbol{w}$ to express $\boldsymbol{x}$ to the user.
\end{itemize}

Ideally, the agent could repeat this cycle to collaborate in real time, as in fig. \ref{fig:rv}.

\paragraph{Example Problem Context for Study}

Agents can give instructions in many settings, for example cooking and equipment assembly. We chose navigation as an example context because it is an old problem with good data and benchmarks~\citep{anderson2018vision}.

Moreover, navigation allows us to directly evaluate our instructions.
Evaluating language is difficult because there is no single standard for good language~\citep{zhao2021evaluation,reiter2009investigation}.
But in navigation, we can compare instructions by giving them to a user and measuring its performance in meters and seconds.
We can then use this evaluation to improve our methods.

Within navigation, we study the ``vision-and-language navigation'' (VLN) task, described in~\citet{anderson2018vision}.
This task uses the Matterport3D environment, which divides real indoor settings into a graph of \textbf{`viewpoints'} spaced about 3m apart~\citep{chang2017matterport3d}.
Each viewpoint has a unique identifier associated with a panoramic RGB-D image taken by a Matterport camera at that location in the environment.
\citet{anderson2018vision} provide a simulator for the environment.

In this environment, VLN uses the
Room-to-Room (R2R) (and similar) datasets~\citep{anderson2018vision},
in which human `guides' describes trajectories of paths in Matterport3D.
Human `followers' ground these descriptions, trying to recreate the original trajectory. 
The resulting dataset is a collection of trajectories paired with instructions, where each trajectory is a series of viewpoints in Matterport3D,
$\boldsymbol{x} = (v_s, \dots, v_g)$, and instructions are series of words, $\underbar{$\boldsymbol{w}$} = (w_0, \dots, w_n)$.

We can now define each language skill in context, summarizing the current literature and opportunities for new research.

\paragraph{Location}

The user begins at a \textbf{start viewpoint} $v_s$, describing this and its \textbf{goal viewpoint} $v_g$  with language $\boldsymbol{l}$ to the \textbf{locator}.
The locator must infer 
$\hat{v}_s$ and $\hat{v}_g$
from $\boldsymbol{l}$.

Studying a similar problem, \citet{tse2018human} use a
bag-of-words model and
a Bayes filter to
track a belief distribution of the agent's location. 
This is a fast and simple method, but it cannot generalize to unseen environments,
because it relies on language occurring alongside locations during training to estimate likelihoods.
Moreover, the bag-of-words approach discards useful information about word order.
\citet{banerjee2020robotslang} created a dataset for a navigation task involving location and grounding.
They give a benchmark for the grounding task, but do not address location.
This illustrates the gap we identify in this paper -- language grounding is well studied at the cost of other skills.

\paragraph{Planning}

Given the locator's guesses of $\hat{v}_s$ and $\hat{v}_g$, the \textbf{planner} 
finds a
feasible path between the estimated start-goal pair,
a \textbf{trajectory} $\boldsymbol{x} = (\hat{v}_s, \dots, \hat{v}_g)$.

In navigation, we have excellent search strategies, so it is tempting to treat this problem as solved.
In our baselines, we use A* search.
A more sophisticated planner, however, would take its user into account.
An optimal path for one user might be a poor choice for another. 
The user of a wheelchair should not be told to take the stairs, for example.
The planner could also consider how well it can describe a path, choosing a simple suboptimal path rather than risk confusion with a shorter one.

We might term this `user-aware planning'.
This is an understudied concept, at least within navigation, and we highlight it as an opportunity for growth.

\paragraph{Generation}

The \textbf{generator} writes \textbf{instructions} $\boldsymbol{w}$ to describe $\boldsymbol{x}$. In Matterport3D, each viewpoint $v_i$ is associated with a panoramic image $p_i$, which can be used for generation.

Because the language is conditioned on a trajectory, we cannot directly use unconditional probabilistic language models
such as GPT-3~\citep{DBLP:brown2020gpt3}.
Rather, our problem is more similar to another conditional generation task, image and video captioning~\citep{you2016image}.
But unlike captions, instructions must come in order and might be quite detailed.

To solve this problem, a common method is to fill templates, such as ``Turn [blank] at the [blank]''.
\citet{daniele2017navigational} used templates in a
$3$-D navigation task, chunking trajectories into series of ``Compound Action Specifictions''.
These specifications reduce instructions to a pattern of orientation (which way to turn) and distance (how far to walk).
Their templates were effective, but the method required engineering features by hand for that environment.
Templates are also inflexible; an instruction might need details like ``Remember that people in Britain drive on the other side."

For the VLN task, \citet{zhao2021evaluation} gives an excellent summary of generation in this context.
They give standards for instructions and evaluate several generators, including their own template generator model, Crafty.
They found generators like the sequence-to-sequence Speaker-Follower~\citep{fried2018speaker} to be relatively successful.
However, VLN researchers mainly use their generators to bootstrap synthetic data and train better grounding models.
This is a great application of a generator, but so far there is less interest in generating instructions for human users: there are many grounding models for VLN, but few generators, and those that do exist are far from human ability~\citep{vlnleaderboard,zhao2021evaluation}.

\section{Data and Standards for Evaluation}
\label{sec:evaluation}

We suggest some simple means to evaluate an agent's ability to give instructions.
Where possible, we use data and metrics from the VLN community.
For the generation task, we use the unmodified R2R dataset, and test on its \textbf{validation seen} and \textbf{unseen} splits~\citep{anderson2018vision}.

The R2R dataset cannot be used directly to train a locating agent,
because it does not associate language to locations.
Instead, we started with the
Room-Across-Room (RxR)~\citep{ku2020room},
which is similiar but includes pose traces and times for each word \textbf{$w_i$} in \textbf{$\boldsymbol{w}$}.
We split up each instruction, 
matching each phrase of \textbf{$w_0 \dots w_j$} to its closest graph location \textbf{$v$},
to make our language dataset.

\begin{figure*}
\begin{center}
\includegraphics[width=\linewidth]{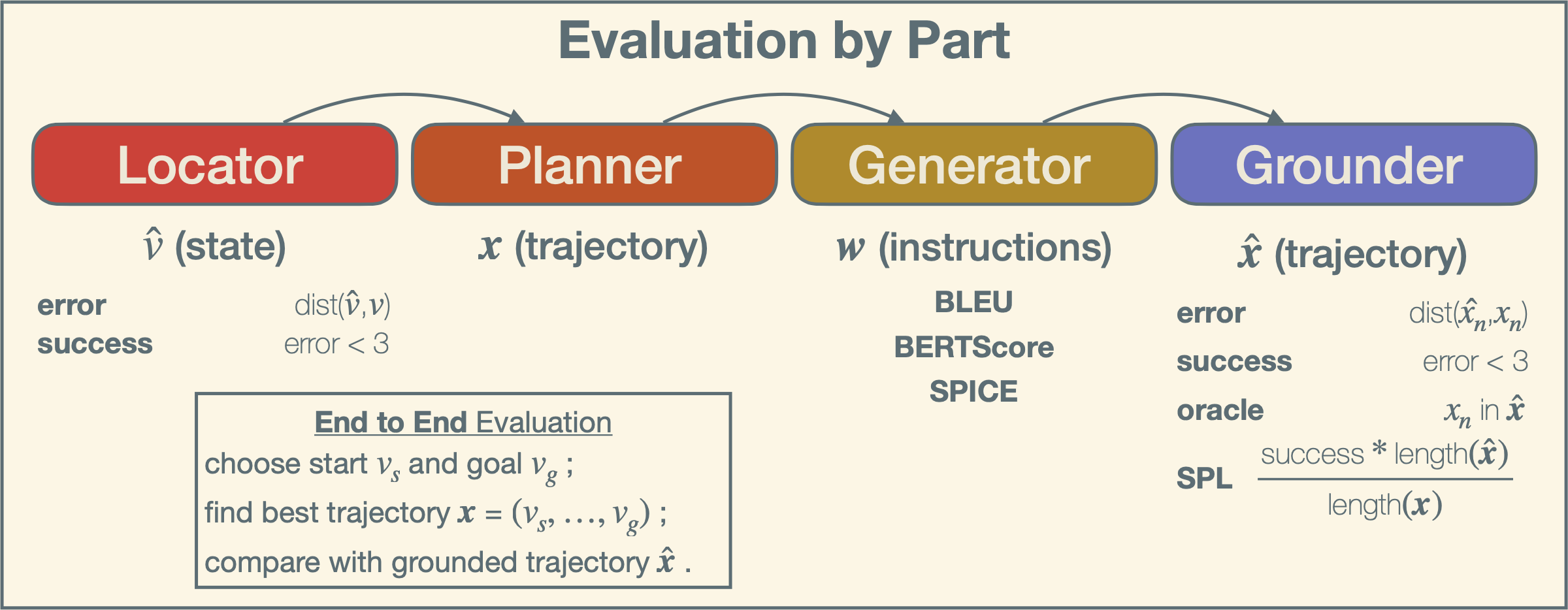}
\caption{Notation and metrics for evaluating tasks individually and as a pipeline.}
\label{fig:eval}
\end{center}
\end{figure*}

\paragraph{Location Evaluation}

We evaluate the locator on its \textbf{error}, the graph distance within the Matterport simulator between its viewpoint $\hat{v}$ and the target viewpoint $v$.
We also report \textbf{success rate};
the task is a success if the error falls within a $3$m threshold,
a convention for this dataset~\citep{anderson2018evaluation}.

\paragraph{Planning Evaluation}

In the sparse graph of the Matterport environment, there tends to be one clearly best plan for any two points, so evaluating the planner made little sense here.
In other environments, we could evaluate planned trajectories by length, elevation change, etc.
We could also test `user-aware' planners by giving them two arbitrary locations
and running their resulting plan through a fixed generator and grounder.
Different planners would elicit different performance from the generator and grounder,
giving us a basis for comparison.

\paragraph{Generation Evaluation}

Generators are usually evaluated by comparing their output to reference translations using metrics like BLEU~\citep{papineni2002bleu} and BERTScore~\citep{zhang2019bertscore}.
There are also models like SPICE~\citep{anderson2016spice}, which evaluate image captions.
These measures were not meant to evaluate instructions
\textit{per se}, and are often of limited value~\citep{zhao2021evaluation}.

\citet{zhao2021evaluation} developed a neural network `compatibility model' to assign instructions and trajectories a similarity score, which correlated well with human evaluation. We do not use their model here; instead, we evaluate our generator directly on how well the user grounds its instructions, producing \textbf{grounded trajectory}
$\hat{\boldsymbol{x}} = (\hat{x}_0, \dots, \hat{x}_n)$.

As in location, we compare the final position of the grounded trajectory $\hat{x}_n$ with the final position of the actual trajectory $x_n$ to find \textbf{error}.
In addition to \textbf{success rate}, there is the more generous \textbf{oracle success rate},
measuring whether the agent \textbf{ever} encountered the goal.
Finally, there is success weighted by path length, \textbf{SPL}, 
which penalizes agents for taking a longer path than necessary~\citep{anderson2018evaluation}.

The R2R dataset includes human descriptions of trajectories $\boldsymbol{x}$; we denote these ground-truth descriptions as \underbar{$\boldsymbol{w}$}.
By using a fixed model to ground first those human instructions, then our generator's description of the same $\boldsymbol{x}$,
we get a sense of how close our generator comes to human ability.
We use a single pretrained grounding model, the Speaker-Follower model~\citep{fried2018speaker}, as the fixed grounder. It is no longer state of the art, but is easily implemented, and good enough for a reasonable comparison of a generator. A better evaluation would include an ensemble of different grounding models, as well as a human evaluation on the simulator.

\paragraph{End-to-End Evaluation}

As location, planning, and generation are parts of a single process,
we suggest an end-to-end evaluation as depicted in 
fig. \ref{fig:eval}.
This allows us to test an agent's performance in a closed loop, 
ideally with a human user.
We tested our baseline models end-to-end as shown, but they performed worse than random agents, so we omit the results table here.

\begin{table*}[h!]
\centering
\caption{Baseline results for generation. Navigation metrics obtained by grounding generated language using a fixed grounding model (see "Data and Standards for Evaluation"). Human-level success rates are $\approx 86\%$.}
\begin{tabular}{c||c|c|c|c|c|c|c|c}
     & \multicolumn{2}{c|}{Error (m) $\downarrow$} & \multicolumn{2}{c|}{Success  (\%) $\uparrow$} & \multicolumn{2}{c|}{Oracle Succ. (\%) $\uparrow$} & \multicolumn{2}{c}{SPL  (\%) $\uparrow$} \\
     & seen & unseen & seen & unseen & seen & unseen & seen & unseen \\
    \hline 
     Random Grounder & 9.66 & 9.46 & 17.05 & 19.41& 22.35 & 22.47 & 16.07 & 16.92 \\
     \hline
     Human Instructions & 3.33 & 6.64 & 66.70 & 35.20 & 74.05 & 44.70 & 60.69 & 28.13 \\
     Seq2Seq & 3.58 & 6.68 & 64.11 & 33.84 & 71.76 & 44.40 & 59.18 & 27.04 \\
     \hline
     Reinforce (BLEU) & \textbf{3.42} & 6.61 & \textbf{65.00} & 32.31 & \textbf{72.35} & 44.32 & \textbf{59.24} & 25.80 \\
     Reinforce (BERT) & 3.69 & 6.62 & 64.68 & 35.52 & 69.71 & 44.70 & 58.42 & 28.23 \\
     Reinforce (Agent) & 3.87 & \textbf{6.38} & 59.90  & \textbf{36.44} & 69.41 & \textbf{47.13} & 54.90 & \textbf{29.18}
\end{tabular}
\end{table*}

\section{Baseline Results and Discussion}

We report empirical results on location and generation using some baseline models, highlighting the large performance gap between existing models and ideal.

\paragraph{Location}

\begin{itemize}[leftmargin=*]
    \item \textbf{Bag of Words:} Empirically estimates $p(v_i | \boldsymbol{l}$) from word frequency in the training text, then greedily chooses from among all viewpoints.
    By design, cannot generalize to unseen viewpoints.
    \item \textbf{RNN/ResNet:} Encodes image $p_i$ at viewpoint $v_i$ with a ResNet~\citep{he2016deep} trained on ImageNet~\citep{deng2009imagenet}, producing $h_v$. Encodes word embeddings of $\boldsymbol{l}$ with an LSTM~\citep{hochreiter1997long} to make a context vector $h_w$. Trains a deep neural network to produce $y=f(h_v,h_w)$, choosing the highest $y$.
    \item \textbf{Stacked Cross Attention:} SCAN~\citep{lee2018stacked} uses attention to align parts of the viewpoint image $p$ with individual words $l \in \boldsymbol{l}$, giving an overall similarity score. Chooses the most similar viewpoint.
\end{itemize}

\begin{table}[h!]
\centering
\caption{Baseline results for location. Success rates are too low to locate users using language.}
\begin{tabular}{c||c|c|c|c}
     & \multicolumn{2}{c}{Error (m) $\downarrow$} & \multicolumn{2}{|c}{Success (\%) $\uparrow$} \\
     & seen & unseen & seen & unseen \\
    \hline 
     Random & 9.58 & 8.31 & 6.56 & 8.00  \\
     \hline
     Bag of Words & 29.14 & - & \textbf{8.08} & - \\
     RNN/ResNet & 8.81 & 8.13 & 19.85 & 20.71 \\
     SCAN & 8.06 & 7.93 & \textbf{29.82} & \textbf{27.34} 
\end{tabular}
\end{table}

\paragraph{Generation}

\begin{itemize}[leftmargin=*]
    \item \textbf{Sequence to Sequence:} A sequence-to-sequence LSTM with attention \citep{fried2018speaker}. Minimizes cross-entropy loss of output tokens,
    $\boldsymbol{w}$ against target tokens, \underbar{$\boldsymbol{w}$}.
    \item \textbf{REINFORCE:} Starting with the pretrained generator above, uses the REINFORCE method~\citep{williams1992simple} to backpropagate an undifferentiable loss to the model. We test two losses based on the standard language metrics BLEU and BERTScore.
    A third loss (Agent) explores the idea of training a generator directly to produce better instructions. We use a pretrained grounding model to take actions in the simulator according to the generated instructions, and return loss equal to the distance between the goal location $v_g$ and the agent's final location
    $\hat{x}_n$.
\end{itemize}

\paragraph{Discussion}
Our baseline results show that there is ample room for improvement across tasks. We close with a few points to summarize and extend our work.

To train our locator, we had to do surgery on our language dataset.
Since the language skills we describe are understudied in robotics,
we must augment existing data,
as well as collect new data.

Although we spent little time on planning,
it is part of the collaborative process and should be studied alongside it.
In planning, the instruction agent is not only choosing the best path for navigation,
it is choosing the best path to tell the user.
Capturing user outcomes in a cost function will help us train better planners.

Finally, we note that our tasks of location, planning, and generation are closely tied to grounding.
Generators can make better training data for grounding,
but grounding is helpful in training generators,
by providing an objective evaluation mechanism for scoring generated language.

This is good reason to build robots that understand both sides of the communication process, opening new opportunities for human-robot collaboration.

\bibliography{ref}

\end{document}